# Earthquake Impact Analysis Based on Text Mining and Social Media Analytics

***Zhe Zheng, Hong-Zheng Shi, Yu-Cheng Zhou, Xin-Zheng Lu, and Jia-Rui Lin****
*Department of Civil Engineering, Tsinghua University, Beijing, China, 100084*

***ABSTRACT:*** *Earthquakes have a deep impact on wide areas, and emergency rescue operations may benefit from social media information about the scope and extent of the disaster. Therefore, this work presents a text mining-based approach to collect and analyze social media data for early earthquake impact analysis. First, disaster-related microblogs are collected from the Sina microblog based on crawler technology. Then, after data cleaning a series of analyses are conducted including (1) the hot words analysis, (2) the trend of the number of microblogs, (3) the trend of public opinion sentiment, and (4) a keyword and rule-based text classification for earthquake impact analysis. Finally, two recent earthquakes with the same magnitude and focal depth in China are analyzed to compare their impacts. The results show that the public opinion trend analysis and the trend of public opinion sentiment can estimate the earthquake's social impact at an early stage, which will be helpful to decision-making and rescue management.*

## 1. Introduction

Earthquakes are potential threats to modern cities with large and concentrated populations and complex networked infrastructure systems (Lu et al., 2020). Extreme earthquakes (e.g., the 2008 Wenchuan earthquake, 2010 Haiti earthquake, and 2011 Tōhoku earthquake) will inflict severe damage and lead to extensive casualties, and significant losses (Cui et al., 2008; DesRoches et al., 2011; Mori et al., 2011). So, time-sensitive responses such as emergency rescue operations should be taken to help people locate available resources, deliver assistance, and so on (Yin et al., 2015). The governments' time-sensitive responses rely on the timely acquisition and analysis of disaster-related information.

On the one hand, with the popularity of the mobile Internet and the increasing use of social media platforms, users are generating massive amounts of data and information anytime, anywhere. The contents generated by the users are a source of big data that could be utilized in decision support systems to help governments and citizens manage disaster risks in terms of vulnerability assessment, early warning, monitoring, and evaluation (Ford et al., 2016). On the other hand, with the advent of natural language processing (NLP) techniques, the analysis and processing of texts are becoming more and more feasible and convenient. For example, more NLP-based applications have emerged, such as the domain-specific language model (Zheng et al., 2022), semantic web (Wu et al., 2022), automated rule checking (ARC) system (Zhou et al., 2022; Zheng et al., 2022), and topic modeling (Lin et al., 2020).

Twitter is one of the most popular social media platforms that caught much attention of researchers (Yao et al., 2021). To date, many seismic analyses have been carried out based on the big data retrieval from Twitter. For example, Sakaki et al. (2012) proposed a real-time earthquake monitoring method that can effectively detect earthquakes at an early stage merely by monitoring tweets. Van Quan et al. (2017) proposed a convolution neural network (CNN) based method to classify informative tweets and the real-time event detection algorithm to detect the occurrence of a certain event. Avvenuti et al. (2018) proposed an emergency management system called CrisMap to visualize the loss of an earthquake based on natural language processing and geoparsing tools. Sina microblog is the most popular social platform in China (Chen et al., 2020), and the big data from Sina have also been utilized in the extraction, detection, and analysis of earthquakes in recent years. Qu et al.（2011）investigated how Chinese netizens used microblogging in response to the 2010 Yushu Earthquake. Xing et al. (2019) proposed an emergency information spatial distribution detection method based on microblog information urgency grading and spatial autocorrelation analysis. Yao et al. (2021) proposed an intensity extraction method for Sina texts. Although many studies have been conducted for earthquake disaster analysis based on Twitter or Sina microblogs, more case studies are still needed to further validate the effectiveness of different earthquake hazard analysis methods.

To explore and estimate the social impacts and perceptions of earthquakes through NLP-based approaches, four text-mining methods (i.e., word frequency analysis, public opinion trend analysis, sentiment analysis, and keyword-based text classification) are introduced first in Section 2. And then two earthquake cases with the same magnitude are used for analysis and validation in Section 3. The results show that the public opinion trend analysis and the sentiment analysis method can estimate the earthquake's impact at an early stage. Finally, Section 4 concludes the findings and limitations of this research.

# 2. Methodology

The earthquake-related microblogs from the Sina microblog are collected utilizing clawers as the data source. Subsequently, data cleaning and preprocessing are performed, including irrelevant data filtering and word segmentation. Then, several NLP-based analyses, such as public opinion trend, sentiment, word frequency, and keyword-based text classification, are conducted to evaluate the earthquake impact. At last, the relationship between post-earthquake public opinion evolution characteristics and earthquake impact is summarized. Fig. 1 shows the outline and workflow of this research. Meanwhile, the toolkits used to implement the workflow are also displayed at the bottom of Fig. 1. The adopted methods for data analysis are detailly illustrated from Section 2.2.1 to Section 2.2.4.

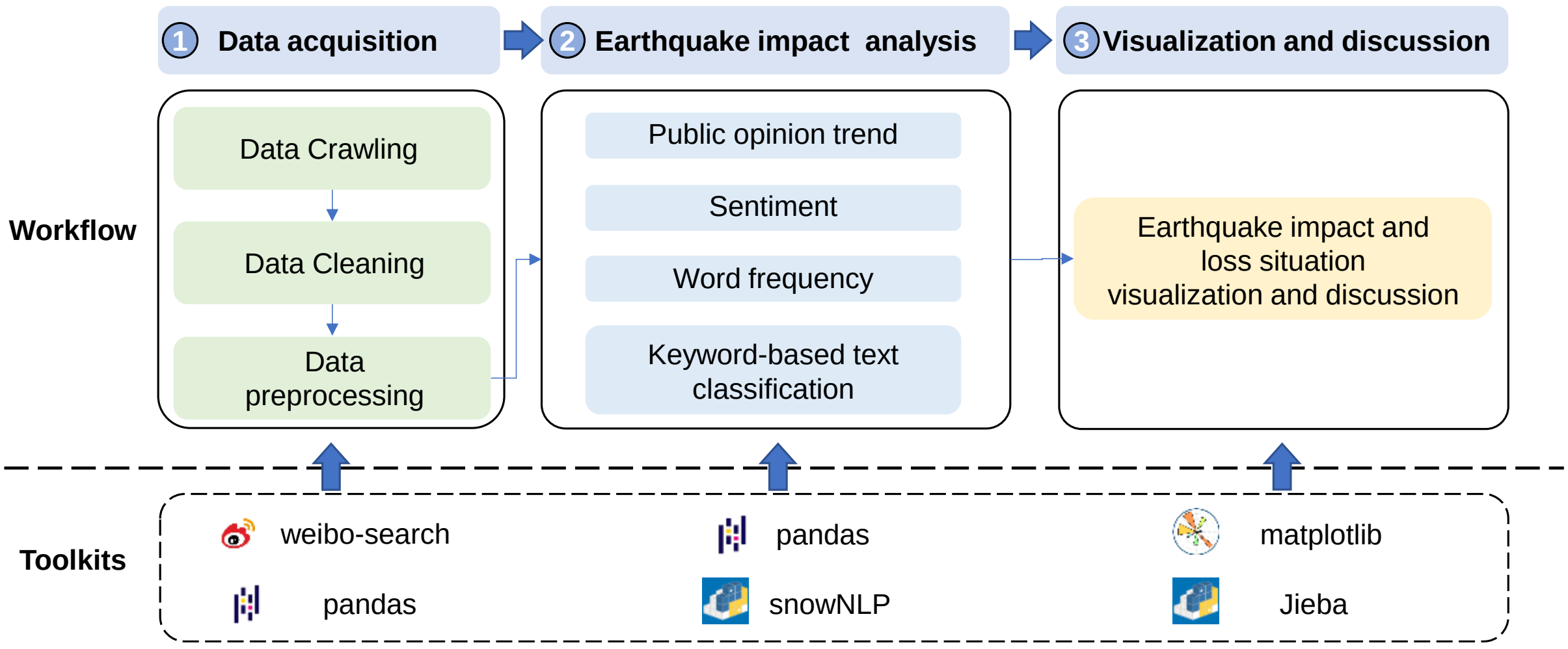

**Fig.1.** Workflow of this research

## 2.1 Data acquisition and preprocessing

The open-source tool named weibo-search is used to acquire raw data. This tool can automatically retrieve and save public information from Sina microblogs, including the full text of microblogs, release time, and publisher based on various restrictions such as keywords, release time, and release location.

There are some irrelevant microblogs in the raw data that do not meet the restrictions. Therefore, further data cleaning is performed, and only the microblogs containing the target keywords and with the proper release time are retained.

In order to facilitate the subsequent analyses, the texts of the microblogs are further preprocessed, including the removal of special symbols, word segmentation, and removal of stop words. The removal of special symbols is implemented using regular expressions, and the removal of word segmentation and stop words is performed based on the jieba package (an open-source Chinese segmentation application in Python language).

## 2.2 Data analysis

### 2.2.1 Word frequency analysis

The word frequency analysis can find the hot spots and characteristics of an earthquake by counting the keywords in the microblogs and the corresponding comments. Based on the preprocessing (Section 2.1), this section further screens out some keywords that only reflect the spatial and temporal information of earthquakes (such as earthquake location, and time). Because the number of these keywords is the largest but the information is limited.

Then the wordcloud Python package is utilized to visualize the word frequency in the form of a word cloud. Finally, the characteristics of the earthquakes reflected in the keywords are analyzed in Section 3.2.1.

### 2.2.2 Public opinion trend analysis

Public opinion trend analysis aims to estimate the impact and scope of an earthquake by counting the number of microblogs released in each time period and analyzing the trend of microblogs. This work obtains the releasing time of each microblog in a unified format (YYYY-MM-DD HH:MM) based on weibo-search. Then, using pandas and matplotlib, the trend of public opinion within 48 hours after the earthquake can be visualized on an hourly basis. At last, the characteristics of the public opinion trend after the earthquake are summarized in Section 3.2.2.

### 2.2.3 Sentiment analysis

Sentiment analysis is to classify whether the emotion of each microblog is positive or negative. Sentiment analysis of earthquake-related microblogs can estimate the economic losses and social impacts caused by an earthquake. Negative emotions may increase when earthquake damage is severe. Typical positive and negative microblogs are shown in Table 1.

This study employs the sentiment analysis model in snowNLP (a Python library for Chinese NLP tasks) to estimate the sentiment of each microblog. Microblogs with a positive emotion probability greater than 50% are considered positive ones, and vice versa. On this basis, the evolution of the number and proportion of two types of microblogs can be calculated and visualized. Finally, the relationships among emotional evolution characteristics, the earthquake impact, and the post-earthquake disposal are discussed in Section 3.2.3.

Table 1. Typical microblogs with two different sentiments

| Sentiment | Text |
|---|---|
| Positive | On April 6, a class of a middle school in Xingwen County, Yibin, Sichuan was reading early when an earthquake warning sounded. The students quickly hunched over their heads and squatted to avoid it, and then began to evacuate in an orderly manner. The students did not panic, and there were no casualties.<br><br>When the earthquake comes, the cat in the house sleeps soundly. |
| Negative | In Xingwen County, the Qishupo section and the Xinhuachang section of the provincial highway (S444) collapsed, causing road interruptions. At present, traffic control has been implemented on the above-mentioned road sections, and vehicles and pedestrians are prohibited from passing through!<br><br>I thought it was my illusion, but there was a real earthquake, and the magnitude was not low. |

### 2.2.4 Keyword-based text classification for disaster evaluation

Keyword-based text classification can estimate the impact of an earthquake more fine-grained. This work employs the keyword list of different disaster levels provided by Kong et al. (Kong et al., 2020), as shown in Table 2. The keyword list contains three levels of keywords: slight, moderate, and severe.

The keyword matching method is used to classify the microblogs. When one microblog contains a slight, moderate, and severe level keyword, the microblog will be classified in that category, respectively. When one microblog contains multiple keywords in multiple types, the classification priority is severe > moderate > slight. After classification, the distribution of disaster-level of earthquake-related microblogs can be analyzed.

Table 2. Keywords for different disaster level

| Disaster level | Keywords |
|---|---|
| Slight | Shake, Loud, Slight, Vibrate, Trapped, Divert, Forbidden, Restricted, Obstructed, Impassable, … |
| Moderate | Obvious, Smashed, Dizzy, Buried, Injured, Fractured, Cracked, Falling Stone, Damaged, Torn, Rolling Stone, … |
| Severe | Ferocious, Fierce, Intense, Severe, Dead, Wrecked, Unfortunate, Life-Threatening, Collapsed, Broken, Landslide, … |

# 3. Case study

## 3.1 Earthquake information and data collection

The Yibin Xingwen earthquake occurred at 7:50 on April 6, 2022, with the epicenter located at 28.22 N and 105.03 W, with a magnitude of 5.1, and a depth of 10 km. Its maximum intensity was 6 degrees, and the 6-degree zone included seven townships, with an area of approximately 379 km$^2$. The population affected within the 6-degree zone was approximately 750,000. The earthquake induced 719 houses damaged, 1,045 residents urgently transferred, 2 provincial roads interrupted, 5 village roads damaged, etc.

The Tangshan Guye earthquake occurred at 6:38 on July 12, 2020, with the epicenter located at 39.78 N and 118.44 W, also with a magnitude of 5.1, and a depth of 10 km. Its maximum intensity was 5 degrees, covering four districts including Guye, Kaiping, Luanzhou, and Qian'an, with an area of approximately 437 km$^2$ and an affected population of about 2.36 million. The earthquake effects only included minor cracks in individual old houses.

For the Xingwen earthquake, the search keyword was "#Sichuan Yibin Xingwen County 5.1 earthquake#" and the release time was limited to April 6, 2022, to May 5, 2022. For the Guye earthquake, due to the lack of hashtags similar to the Xingwen earthquake, the search keywords were set to include both "Guye" and "earthquake", and the release time was limited to July 12, 2020, to July 12, 2021. After data collection and cleaning, the data volume for the Xingwen earthquake was 2081, and the data volume for the Guye earthquake was 7584.

Table 3. Overview of the dataset.

| | Xingwen Earthquake | Guye Earthquake |
|---|---|---|
| Keywords for searching | #Sichuan Yibin Xingwen County 5.1 earthquake# | "Guye" and "earthquake" |
| Time range | 2022/04/06-2022/05/05 | 2020/07/12-2021/07/12 |
| Volume of data | 2081 | 7584 |

## 3.2 Analysis result

### 3.2.1 Word frequency analysis

In this study, word clouds of hot words within 24 hours and 24-48 hours after the two earthquakes are drawn, respectively, as shown in Fig. 2. Through the analysis of the word cloud, several main categories of post-earthquake microblogs can be summarized, such as the basic information of the earthquake (time, depth, location, and so on), the rescue force (fire fighting), and the knowledge of earthquake (risk avoidance, escape methods). Therefore, the word cloud is not very helpful for evaluating the impact of earthquakes, and it is relatively weak in assisting decision-making in the early stage of earthquakes.

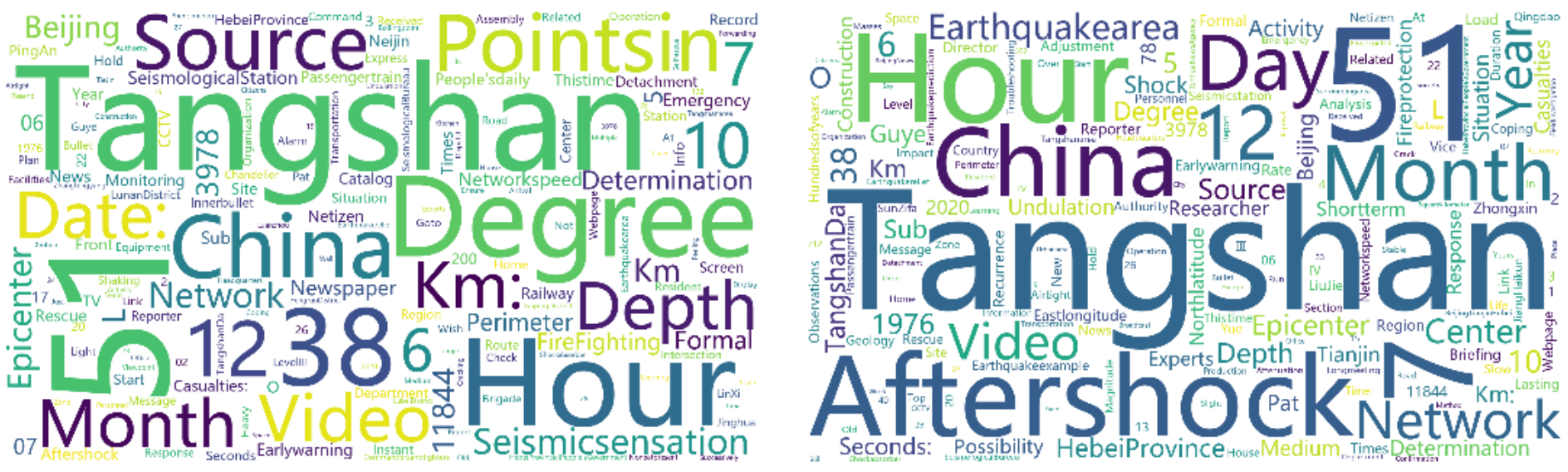

(a) Wordcloud for Guye erthquake (24h) (b) Wordcloud for Guye erthquake (24-48h)

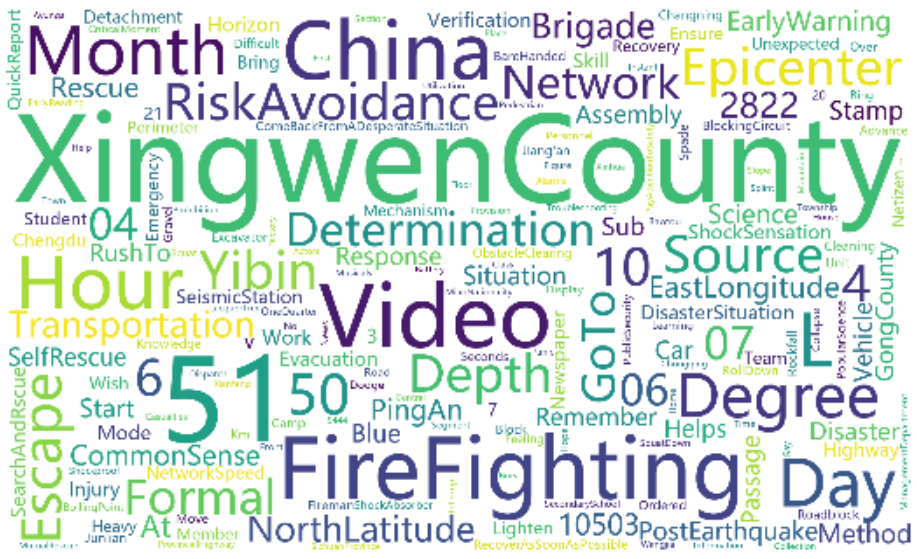

(c) Wordcloud for Xingwen erthquake (24h)

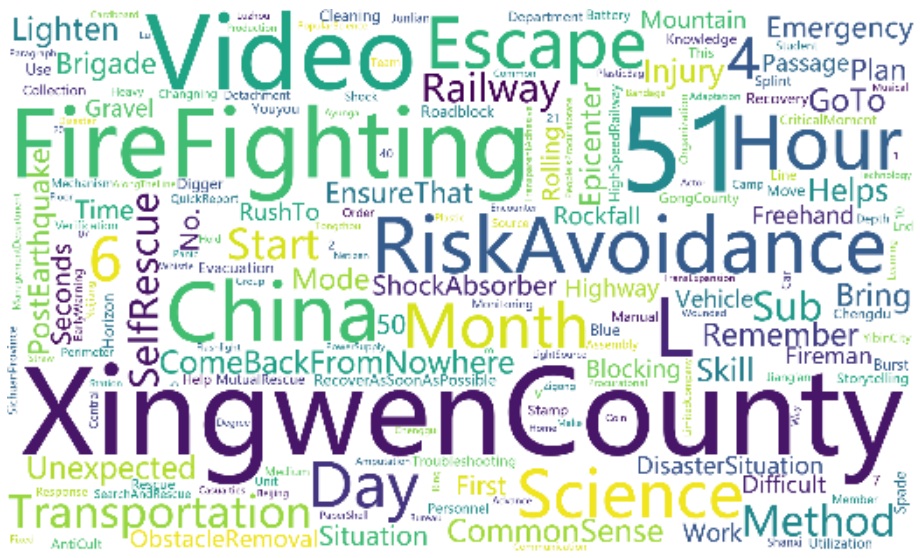

(d) Wordcloud for Xingwen erthquake (24-48h)

Fig. 2. Wordclouds within 24h and 24-48h after the two earthquakes

### 3.2.2 Public opinion trend analysis

Fig. 3 shows the public opinion trend within 48 hours after the two earthquakes. The public opinion of the two earthquakes both peaked at 1 hour after the earthquake, and then declined rapidly. It is worth noting that the evolution tends to end near 48 hours, so the subsequent analysis of this study will mainly be based on microblogs within 48 hours after the earthquake.

In general, the peak public opinion volume can be used to estimate the impact area of this earthquake. But it should be noted that peak volume only relates to the impact area, and relates little to the damage caused by the earthquake. Taking this study as an example, the losses caused by the Xingwen earthquake are relatively high, but the peak volume of the Xingwen earthquake is significantly lower than that of the Guye earthquake. Because the population in the area with the highest intensity of the Xingwen earthquake is only 750,000, which is far lower than the population of approximately 2.36 million in the area with the highest intensity of the Guye earthquake. Besides, the neighboring large cities (i.e., Beijing and Tianjin) also felt the earthquake after the Guye earthquake. Therefore, the Guye earthquake aroused more discussion online.

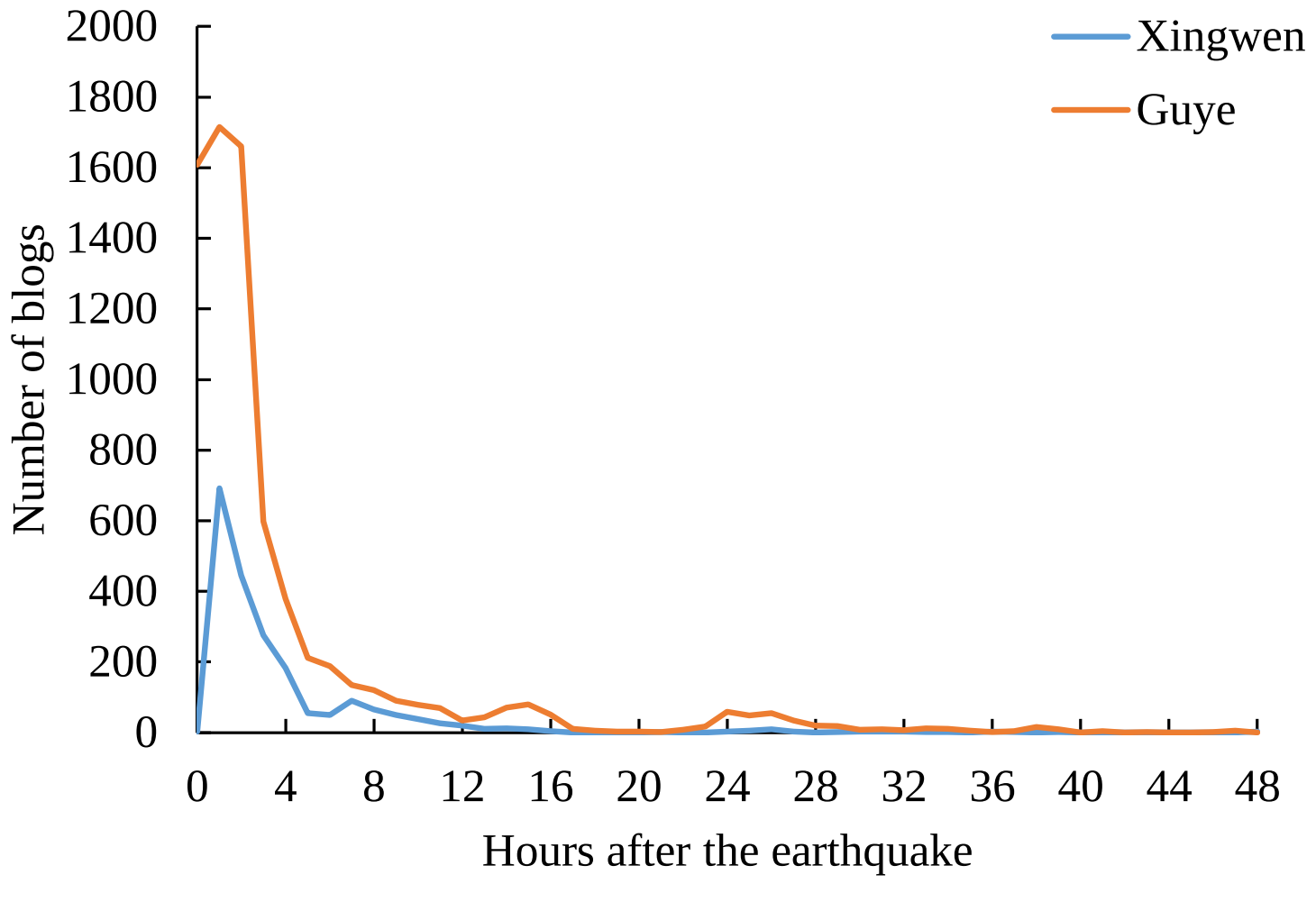

Fig. 3. Public opinion trend of the two earthquakes

### 3.2.3 Sentiment analysis

Figs 4 and 5 show the trends of positive and negative microblogs within 48 hours after the Guye earthquake and after the Xingwen earthquake, respectively. The proportion of negative emotions in microblogs after the Xingwen earthquake was significantly higher than that in the Guye earthquake, which is consistent with the loss situation mentioned in Section 3.1. The contents of microblogs in the early period after an earthquake are mainly spontaneous discussions by people in the epicenter and the affected surrounding areas. Therefore, the high proportion of negative microblogs within 2 hours after an earthquake is more likely that this earthquake is more powerful and with larger losses. In addition, with the progress of disaster relief and rescue, the sentiment of microblogs will be more positive as time goes by. Therefore, the emotional proportion of microblogs can be monitored in real-time after the earthquake to estimate the damage caused by the earthquake to provide a reference for early disaster relief decisions after the earthquake.

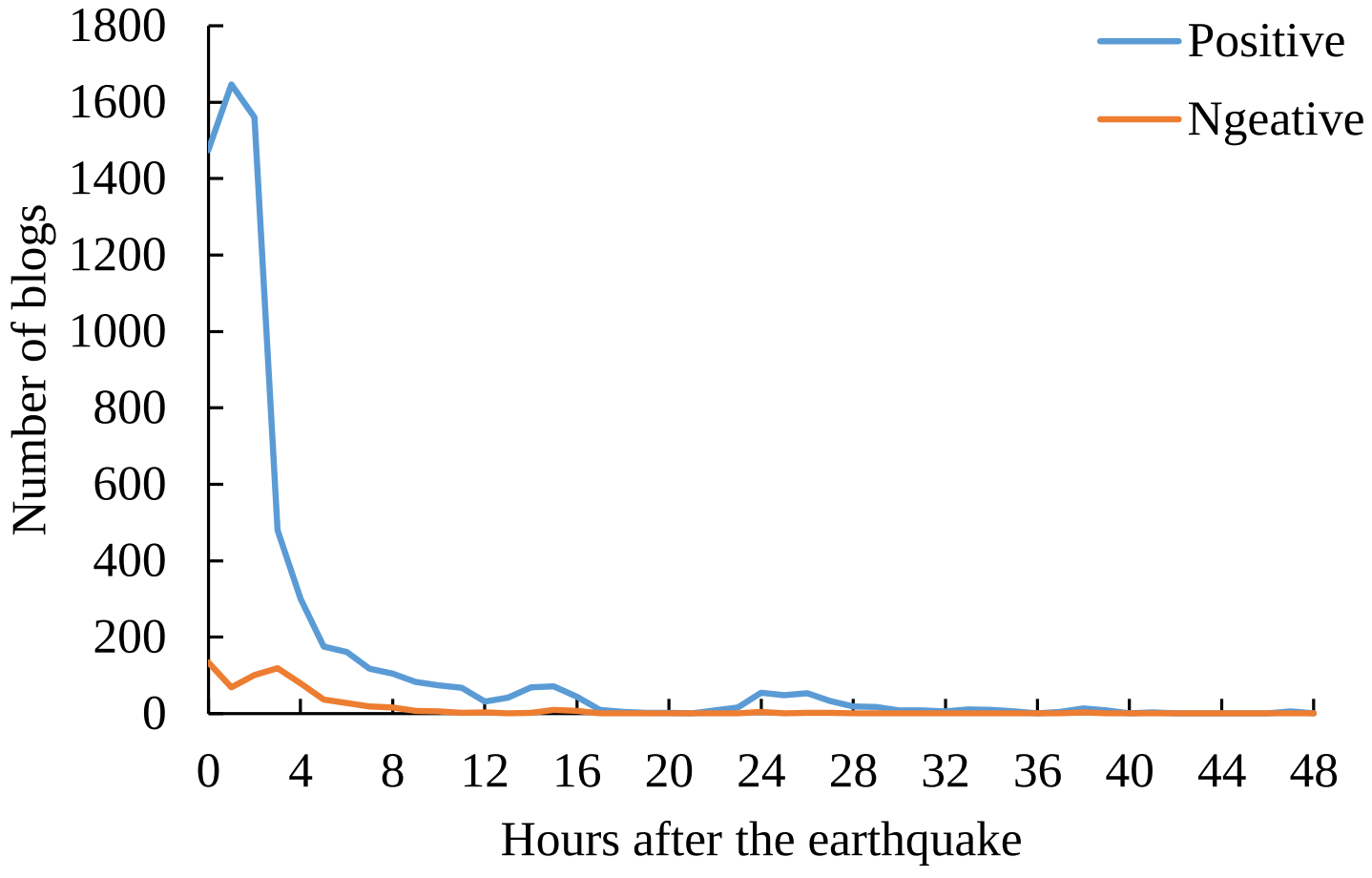

Fig. 4. Sentiment trend after the Guye earthquake

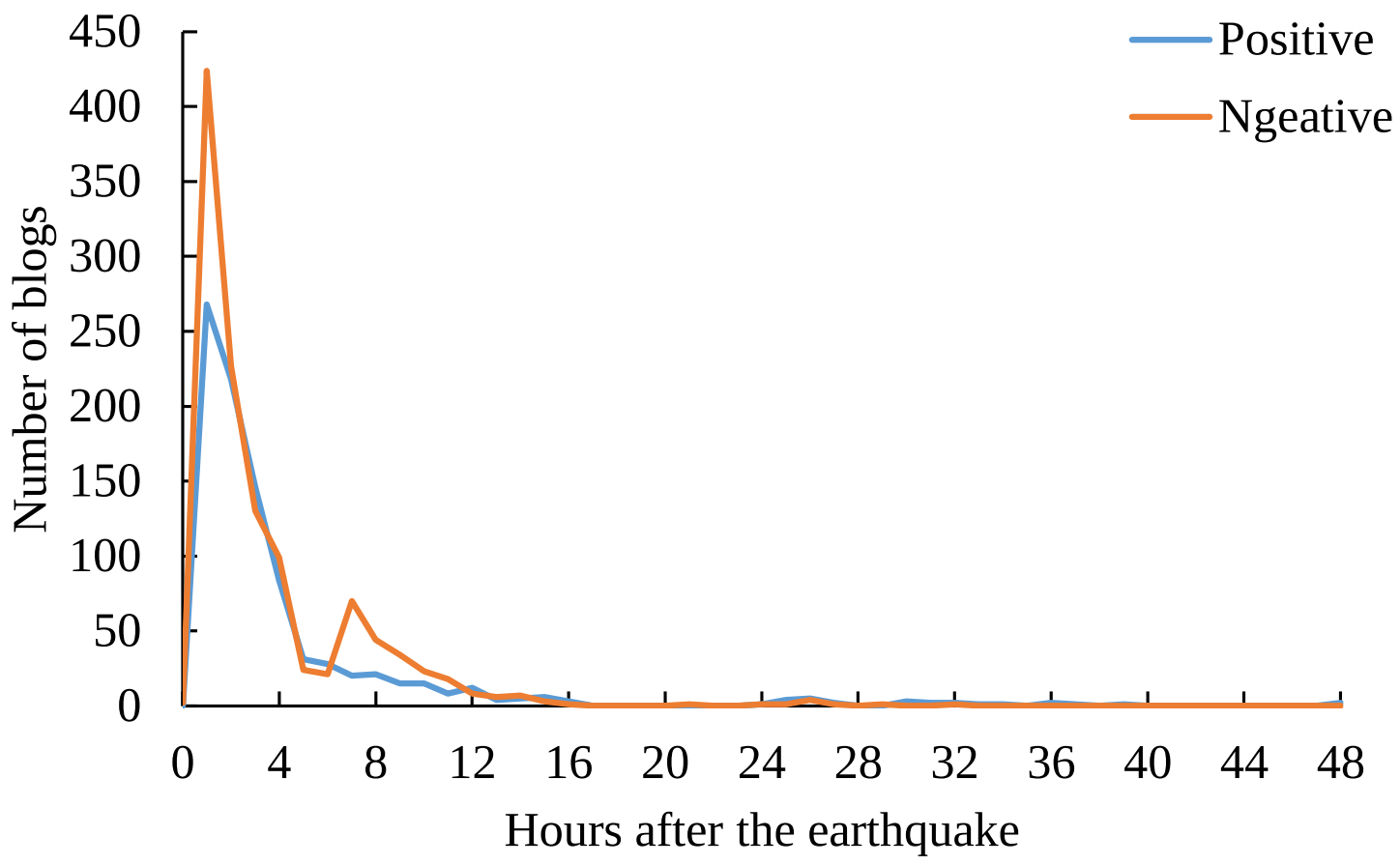

Fig. 5. Sentiment trend after the Xingwen earthquake

### 3.2.4 Keyword-based text classification for disaster evaluation

Using the method described in Section 2.2.4, the disaster levels of each microblog of the two earthquakes were evaluated, and the proportion of each disaster level is shown in Fig. 6. Fig. 6 shows that the impacts of the Guye earthquake are stronger than those of the Xingwen earthquake. The assessment results are weakly correlated with the actual loss situation, as illustrated in Section 3.1. The possible reasons include (1) there are some microblogs about the knowledge of earthquakes (risk avoidance, escape methods) in the existing data. This type of microblog usually describes the severe consequences of an earthquake, thus containing many keywords in Table 2. However, it cannot reflect the specific earthquake loss situation, which will affect the assessment results. Therefore, more accurate seismic-related microblog classification methods (e.g., machine learning-based methods, or deep learning-based methods) should be investigated; (2) whether the selected keywords are appropriate remains to be further investigated.

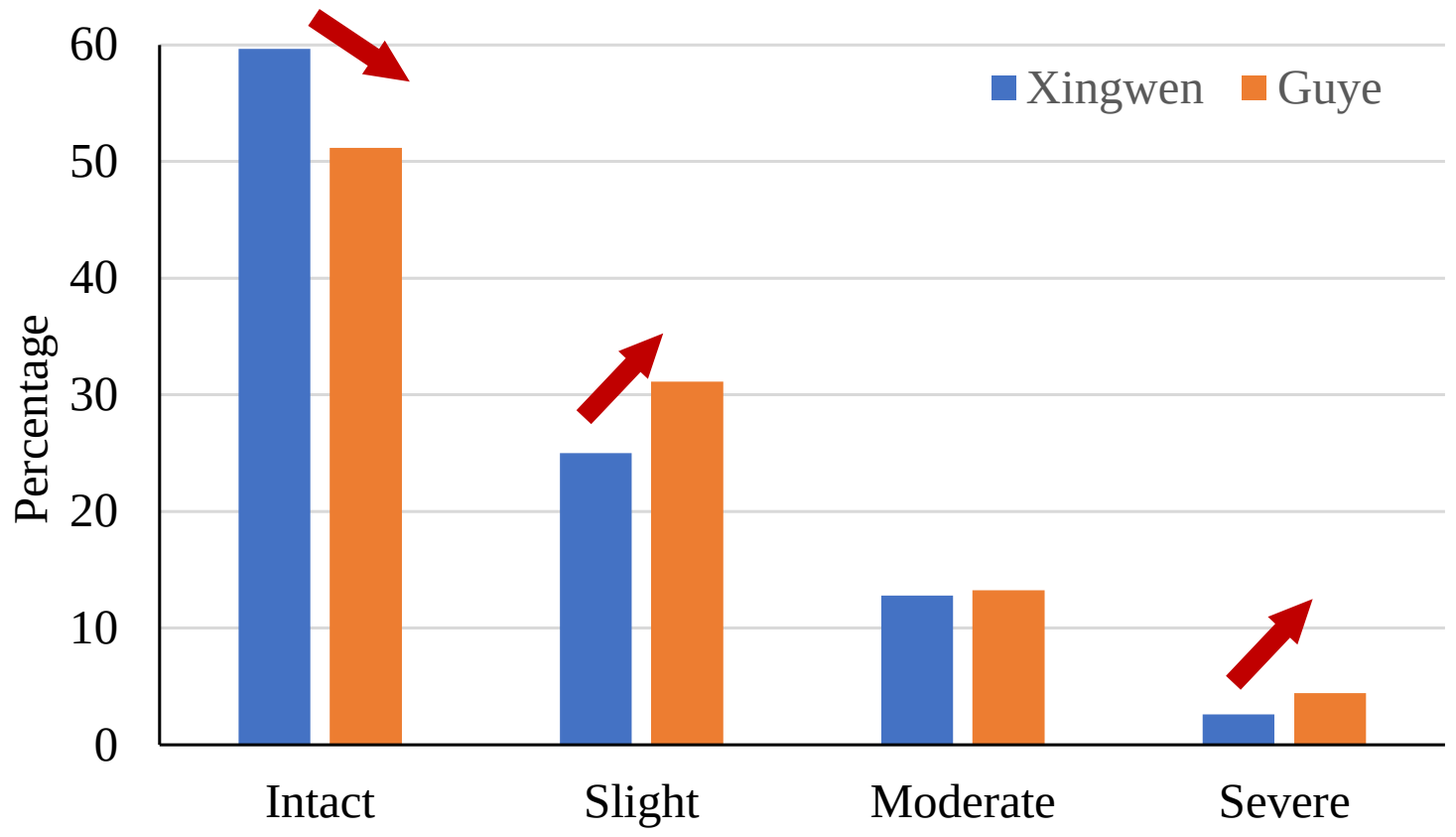

Fig. 6. Percentage of microblogs with different disaster levels

## 4. Conclusion

Earthquakes can have a deep impact on wide areas, and emergency rescue operations can benefit from early information about the scope and extent of the disaster. With the popularity of the mobile Internet, social media has accumulated a large amount of data and information, which can be utilized in decision support systems to help governments for managing seismic-related events in an early stage. To explore the effect of different text mining-based methods in estimating the earthquake's impact, this work established four widely-used methods including 1) word frequency analysis, 2) public opinion trend analysis, 3) sentiment analysis, and 4) keyword and rule-based text classification for earthquake impact analysis. Then two recent earthquakes with the same earthquake magnitude and focal depth in China are analyzed to compare their impact. The conclusions are as follows:

(1) The social impact caused by earthquake disasters is complex. As the cases in this work show, despite similar indicators such as magnitude and depth of the source, the two earthquakes have significantly different social impacts. Therefore, the estimation and perception of the earthquake's social impact at an early stage will be helpful to decision-making and rescue management.
(2) The hot word frequency analysis has little significance for the decision-making in the early stage.
(3) The public opinion volume peaks rapidly after the earthquake and then decreases, and the evolution of public opinion basically tends to end within 48 hours for the two analyzed earthquakes. The peak volume of public opinion can reflect the impact of the earthquake.
(4) The sentiment evolution can reflect the earthquake loss more accurately in the early stage. The more negative microblogs account for the more serious losses.
(5) Keyword-based loss analysis relies on the data cleaning and accurate classification of earthquake loss-related microblogs. Noisy microblogs on social media may have a large influence on its results.

The limitations of this study are listed as follows:

(1) Analysis of more earthquakes is needed to determine the applicability and accuracy of the text mining-based methods.
(2) Sentiment analysis with more fine-grained time, such as accurate to five minutes, should be further explored for a better loss estimation effect.
(3) More accurate seismic-related microblogs classification methods (e.g., machine learning-based methods, or deep learning-based methods) should be investigated.
(4) Further research on the extraction of the geographic location from microblogs should be considered.

## Acknowledgment

The authors are grateful for the financial support received from the National Natural Science Foundation of China (No. 72091512, No. 51908323) and the Tencent Foundation through the XPLORER PRIZE.